\documentclass[12pt]{article}
\usepackage{fullpage}
\usepackage{setspace}
\usepackage{hyperref}
\usepackage{fancyhdr}
\usepackage{lastpage}
\usepackage[pdftex]{graphicx}
\usepackage{amsmath}
\usepackage{listings}
\let\stdsection\section
\renewcommand\section{\newpage\stdsection}

\begin{document}

\title{Language Acquisition in Computers}
\date{April 2012}
\author{Megan Belzner\\\url{belzner@mit.edu}
		\and Sean Colin-Ellerin\\\url{seancolinellerin@gmail.com}
		\and Jorge H. Roman\\\url{jhr@lanl.gov}}

\doublespacing

\maketitle

\begin{abstract}
This project explores the nature of language acquisition in computers, guided by techniques similar to those used in children. While existing natural language processing methods are limited in scope and understanding, our system aims to gain an understanding of language from first principles and hence minimal initial input. The first portion of our system was implemented in Java and is focused on understanding the morphology of language using bigrams. We use frequency distributions and differences between them to define and distinguish languages. English and French texts were analyzed to determine a difference threshold of 55 before the texts are considered to be in different languages, and this threshold was verified using Spanish texts. The second portion of our system focuses on gaining an understanding of the syntax of a language using a recursive method. The program uses one of two possible methods to analyze given sentences based on either sentence patterns or surrounding words. Both methods have been implemented in C++. The program is able to understand the structure of simple sentences and learn new words. In addition, we have provided some suggestions regarding future work and potential extensions of the existing program.
\end{abstract}
 
\begin{singlespacing}
\tableofcontents
\end{singlespacing}
 
\section{Introduction}
 
Natural language processing is a wide and varied subfield of artificial intelligence. The question of how best to give a computer an intuitive understanding of language is one with many possible answers, which nonetheless has not yet been answered satisfactorily. Most existing programs work only within a limited scope, and in most cases it cannot realistically be said that the computer actually understands the language in question.

This project seeks to give a computer a truly intuitive understanding of a given language, by developing methods which allow the computer to learn the language with a minimum of outside input, on its own terms. We have developed methods to teach the computer both the morphology and the syntax of a language, and have provided some suggestions regarding language acquisition at the semantic level.

\subsection{History}

The idea of natural language processing originates from Alan Turing, a British computer scientist, who formulated a hypothetical test, known as the ``Turing Test''.  The ``Turing Test'' proposes that the question ``Can machines think?'' can be answered if a computer is indistinguishable from a human in all facets of thought, such as conversation; object identification based on given properties; and so forth \cite{turing}. After Turing's proposition, many attempts have been made to create natural language processing software, particularly using sound recognition, which is currently used in cell-phones, most proficiently in the iPhone 4S Siri system™.  However, most of these programs do not have high-level semantic abilities, rather they have a very limited set of operations, for which keywords are assigned.  For example, the Siri system™ can send an email or text message.  When told to send an email or text message, the software uses these keywords to open a blank e-mail or text message and when told what is to be written in the e-mail or text message, there is no semantic understanding of the message, simply a transcription of the words using voice recognition \cite{siri}. Similarly, there is a lot of software, such as LingPipe, that is able to determine the origin and basic `significance' of a term or sentence by searching the term(s) on a database for other uses of the term(s).  These programs do not, however, gain a semantic understanding of the term(s), rather they simply collect patterns of information with association to the term(s) \cite{lingpipe}.

\subsubsection{Statistical Parsers}

\begin{figure}[htb]
	\begin{center}
		\vspace{3mm}
		\includegraphics[width=\textwidth]{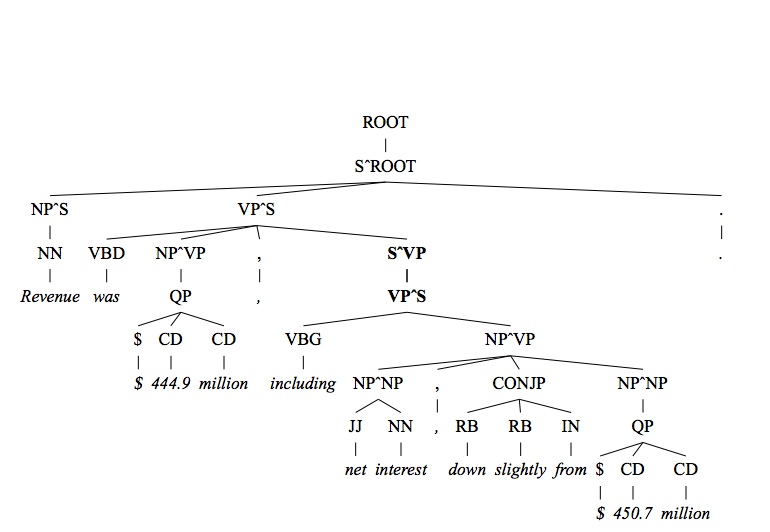}
	\end{center}
	\caption{Statistical parser tree \cite{stanford}} \label{tree}
\end{figure}

There have also been some more technical, less commercially efficacious attempts at natural language processing, such as statistical language parsers, which have been intricately developed by many educational institutions.  Parsers are a set of algorithms that determine the parts of speech of the words in a given sentence.  Current parsers use a set of human-parsed sentences that creates a probability distribution, which is then used as a statistical model for parsing other sentences.  Stanford University and the University of California Berkeley use probabilistic context-free grammars (PCFG) statistical parsers, which are the most accurate statistical parsers currently used, with 86.36\% and 88\% accuracy, respectively \cite{stanford} \cite{berkeley}. The different parts of speech are separated as in Figure \ref{tree}.

In Figure 1, NN $=$ noun, NP $=$ noun phrase, S $=$ subject, VP $=$ verb phrase, and the other symbols represent more specific parts of speech.  One can see that the parser splits the sentence into three parts: the subject noun phrase, the verb phrase, and the noun phrase.  Each of these parts is then split into more parts and those parts into parts, finally arriving at individual qualifications for each word.  The assignment of a given part of speech for a word is determined by tentatively allocating the part of speech that is most probable for that word, which is then tested within its phrase (i.e. subject noun phrase, or verb phrase, etc.), and if the probability remains high then that part of speech is set for the word.  These parsers are called context-free because the parse of a word is not affected by the other words in the sentence other than those in its phrase, while the less accurate parsers obtain an overall probability for a sentence and adjust their parsing accordingly \cite{stats}.

\subsubsection{Other NLP Programs}

In addition to statistical parsers, which only determine the syntax of a sentence, some elementary programs have been written for evaluating the semantics of a given body of text.  There is a system called FRUMP that organises news stories by finding key words in an article that match a set of scripts and then assigns the article to a certain category, in which it is grouped with other articles of similar content.  SCISOR summarizes a given news article by analyzing the events, utilizing three different sets of knowledge: semantic knowledge, abstract knowledge, and event knowledge.  As the program sifts through the body of text, certain words trigger different pieces of knowledge, which are compiled to gain the best understanding of that word or sequence of words.  The resultant meanings can then be organized and rewritten using similar meanings, equally balanced among the three sets of knowledge as the original piece of information.  Similarly, TOPIC summarizes a given text by distinguishing the nouns and noun phrases and then analyzing their meaning through a ``thesaurus-like ontological knowledge base'', after which the program uses these equivalent meanings to rewrite the text \cite{elang}.

\subsection{Common Language Properties}

Certain elements of language are commonly used both in natural language processing and the general analysis of language. These properties include bigrams and recursion, two properties which play a significant role in this project.

\subsubsection{Bigrams}

An $n$-gram is a sequence of $n$ letters.  The most commonly used forms of $n$-grams are bigrams and trigrams because these offer a specific indication for a set of information, without signifying extremely rare and complex aspects of the subject.  A good example of this is the use of bigrams in cryptography.  A common method of decoding a message that has been encoded using a keyword, like the Vigenere Cipher encryption, is to calculate the distance in letters between two of the same bigrams in order to determine the length of the keyword, and then the keyword itself \cite{cipher}. If any n-grams are used, where n is greater than or equal to 4 or even in some cases if n is equal to 3,  then the number of same n-grams for a given message would be very rare and make determining the length of the keyword increasingly difficult.

Similarly, in natural language processing, bigrams are used for word discrimination, which is the understanding of an unknown word based upon bigram correspondence with a reference list of known words.  In addition, word bigrams are used in some lexical parsers, the Markov Model for bag generation, and several text categorization tools \cite{markov}.

\subsubsection{Recursion}

The principle of recursion is an essential aspect of human language, and is considered one of the primary ways in which children learn a language.  For a sentence with a particular pattern, a word with a specific part of speech can be exchanged for another word of the same part of speech, indicating that the two words have the same part of speech.  For example, given the sentence: ``The boy wears a hat'', it can be determined that ``the'' and ``a'' are the same part of speech by reconstructing the sentence as ``A boy wears the hat''.  In addition, the word ``boy'' can be exchanged for the word ``girl'', indicating that these are also the same type of word, thereby expanding the lexicon of the child.

In addition, the words of a sentence can remain unchanged, while the pattern changes, thereby introducing a new part of speech.  If we have the sentence ``The boy wears a hat'' or ``A B C A B'' if represented as a pattern of parts of speech, we can add a ``D'' part of speech by creating a new sentence, ``The boy wears a big hat'' (A B C A D B).  The child ascertains that ``big'' must be a new part of speech because no words have previously been placed between an ``A'' and a ``B''.  This method can be repeated for any new part of speech, as well as embedded clauses such as ``The boy, who is very funny, wears a big hat.''

Finally, recursion can be used to indicate the grammatical structures of a language.  Let us examine the following poem:

\begin{center}
\begin{singlespacing}
When tweetle beetles fight,\\
it's called\\
a tweetle beetle battle.\\
\vspace{7 mm}
And when they\\
battle in a puddle,\\
it's a tweetle\\
beetle puddle battle.\\
\vspace{7 mm}
AND when tweetle beetles\\
battle with paddles in a puddle,\\
they call it a tweetle\\
beetle puddle paddle battle.\\
\end{singlespacing}
\hspace{35 mm} - Excerpt from Dr. Seuss' \emph{Fox in Socks} \cite{poem}
\end{center}

The author uses the recursive principle to indicate that ``tweetle beetle'' can be both a noun and an adjective, and then repeats this demonstration with ``puddle'' and ``paddle''.  Further, the correct placement of the new noun acting as an adjective is shown to be between the old string of nouns acting as adjectives and the object noun.  Conversely, the poem illustrates that a noun acting as an adjective can be rewritten as a preposition and an added clause, e.g ``a tweetle beetle puddle battle'' can be rephrased as ``a tweetle beetle battle in a puddle'' \cite{recursion}.
Thus, the principle of recursion can allow a child to acquire new vocabulary, new types of parts of speech, and new forms of grammar.

\subsection{Linguistic Interpretation}

There is a basic three-link chain in the structure of language. Phonetics is the most basic structure, which is formed into meaning by units, known as words.  Units are then arranged syntactically to form sentences, which in turn forms a more extensive meaning, formally called semantics \cite{child}.

It is fundamental in learning a language that a computer understand the connections in the phonetics-syntax-semantics chain, and the structure and computations of each, with the exception of phonetics.  Phonetics and their formulation can be disregarded because these refer more greatly to the connections of the brain to external sounds, than the core structure of language.  In fact, the entire external world can be ignored, as there need not be an inference between external objects in their human sensory perception, and their representation as language, formally known as pragmatics or in Chomskyan linguistics as E-language (External Language).  Instead, the relations between words and meaning, intricately augmented by the semantics created by their syntactically accurate formed sentences, is the only form of language necessary, denominated contrastingly as I-language (Internal Language) \cite{child}. Noam Chomsky argues that I-Language is the only form of language that can be studied scientifically because it represents the innate structure of language \cite{elang}. Language similar to this form can be found in humans who are literate in a language, yet can not speak it. Therefore, due to the increase in structure, the lack of external representation in natural language processing may be more advantageous than one might think.

\subsection{Project Definition}

This project examines the nature of language acquisition in computers by implementing techniques similar to those used by children to acquire language. We have focused primarily on morphology and syntax, developing methods to allow a computer to gain knowledge of these aspects of language. We have developed programs in both C++ and Java. 

Regarding morphology, the program is able to analyze the word structure of given languages and distinguish between languages in different samples of text using bigram frequencies, and we have examined the usefulness and limitations of this method in the context of existing methods. Using this technique we have developed computationally understandable definitions of English, French and Spanish morphologies. We have also described and partially implemented a novel technique for understanding the syntax of a language using a minimum of initial input and recursive methods of learning both approximate meanings of words and valid sentence structures. Finally, we provide suggestions for future work regarding the further development of our methods for understanding syntax as well as potential methods for gaining a rudimentary understanding of semantics.

\section{Morphology}
To analyze the morphology of a given language, bigrams can be used to define and compare languages. Since the frequency distribution of a set of bigrams is unique to a given language (across a large enough sample text), this can be used as an accurate identifier of a language with minimal effort.

\subsection{Algorithm}
The program was initially developed in C++ then translated to Java to take advantage of non-standard characters, and is set up in two main portions. The first step involves generating a table of frequency values from a file, and the second step is to compare the two tables and determine the level of similarity.

\begin{figure}[htb]
	\begin{center}
		\vspace{3mm}
		\includegraphics[width=\textwidth]{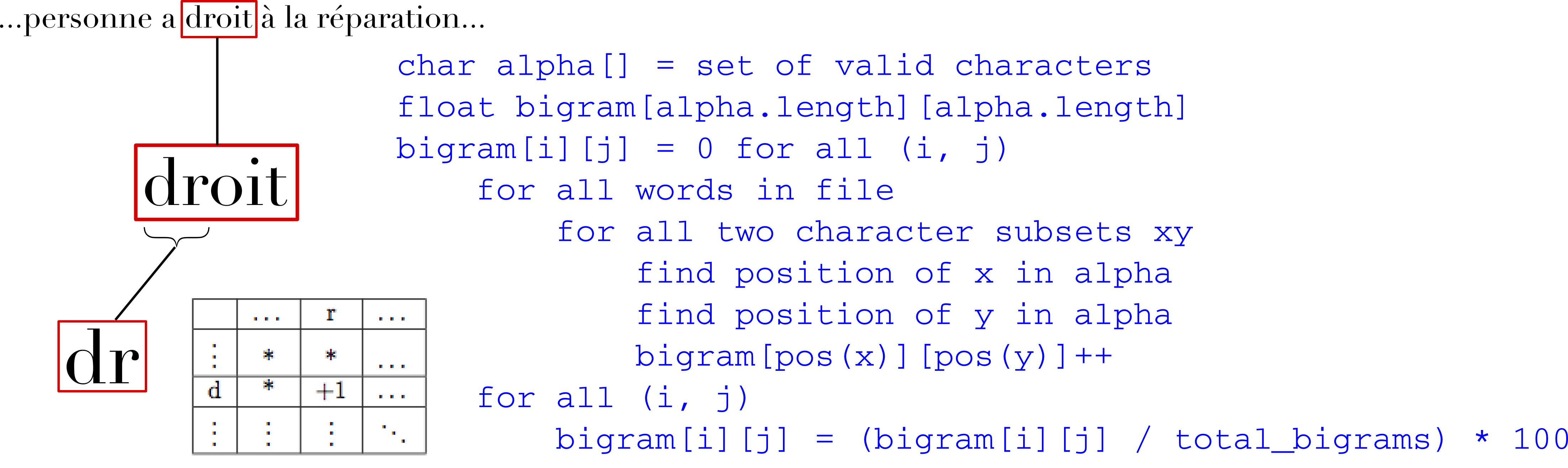}
	\end{center}
	\caption{Learning morphology with bigrams} \label{bigram}
\end{figure}

To generate a frequency table, a two-dimensional numerical array is created from the set of valid characters such that the array includes a space for every possible bigram, with the initial value for each being set to zero. For each word from the input file, the program checks each pair of letters, adding one to the corresponding position in the bigram frequency table. Once the end of the file is reached, each frequency count is divided by the total number of bigrams found times 100 to give a percentage frequency. This process is shown in Figure \ref{bigram}.

This produces an array similar to Table \ref{array}, which can be analyzed separately to examine common and uncommon bigrams for a given language, or compared with another text's table to distinguish between languages as detailed below.

\begin{table}[h]
	\begin{center}
		\vspace{3mm}
		\begin{tabular}{|c|c|c|c|c|} \hline
			 & a & b & c & \ldots \\ \hline
			a & 0 & 0.204 & 0.509 & \ldots \\ \hline
			b & 0.080 & 0 & 0 & \ldots \\ \hline
			c & 0.298 & 0 & 0.269 & \ldots \\ \hline
			$\vdots$ & $\vdots$ & $\vdots$ & $\vdots$ & $\ddots$ \\ \hline
		\end{tabular}
		\caption{Sample of frequency array} \label{array}
	\end{center}
\end{table}

\begin{figure}[htb]
	\begin{center}
		\vspace{3mm}
		\includegraphics[width=0.9\textwidth]{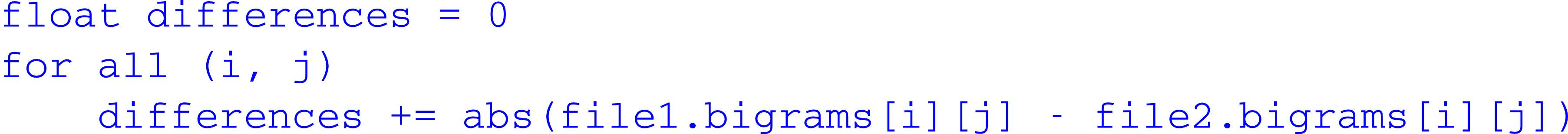}
	\end{center}
	\caption{Calculating morphology differences between sample texts} \label{diff}
\end{figure}

After the frequency tables are created for each file, the two must be compared to determine the level of similarity between the languages of the two files. This is done by finding the absolute values of the differences between corresponding frequencies for the two files, then finding the sum of these differences as seen in Figure \ref{diff}.

This gives an approximate measure of how different the two files are in terms of the frequency of given bigrams. As each language tends to have a unique frequency distribution, a large net difference suggests a different language for each file while a smaller net difference suggests the same language. The threshold dividing a determination of `same language' or `different language' was experimentally determined to be approximately 55.

\subsection{Limitations}

This method does have certain limitations, however. Since the program deals with bigrams (though it can be easily made to use $n$-grams for any $n$ greater than 1), single-letter words are not taken into account. While this does not have a large overall effect, it produces some inaccuracies in the analysis of frequencies for a given language.

A more significant limitation is the requirement that all ``legal'' characters be defined before the program is run. Although it would be relatively straightforward to dynamically determine the character set based on the input files, this creates issues where the character sets for each file are not the same, making it difficult, if not impossible, to accurately compare the two files. Even ignoring this, varying the number of characters may produce variations in the threshold used to determine language similarity. The program is also effective only for files of considerable length to allow for a large enough sample size.

\subsection{Results}

\begin{figure}[h]
	\begin{center}
		\vspace{3mm}
		\includegraphics[width=0.6\textwidth]{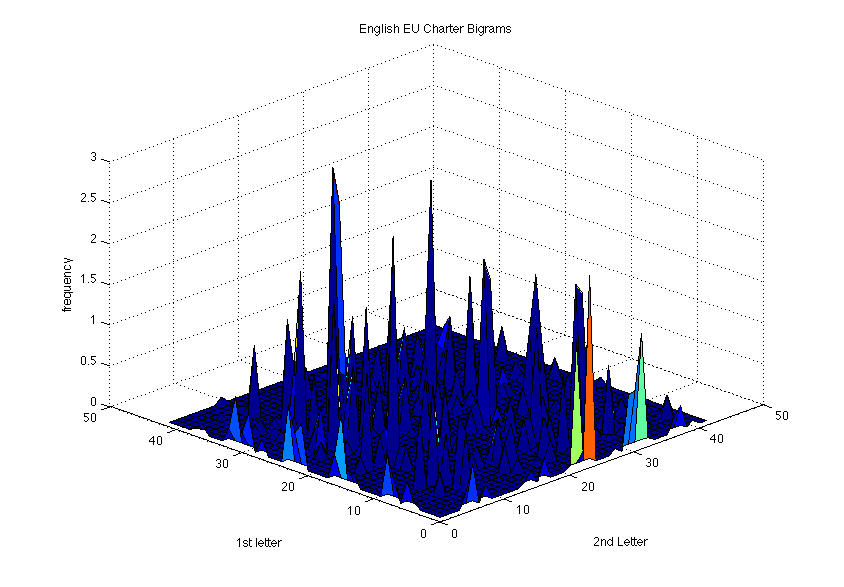}
	\end{center}
	\caption{Distribution of English bigrams} \label{distr_en}
\end{figure}

\begin{figure}[h]
	\begin{center}
		\vspace{3mm}
		\includegraphics[width=0.6\textwidth]{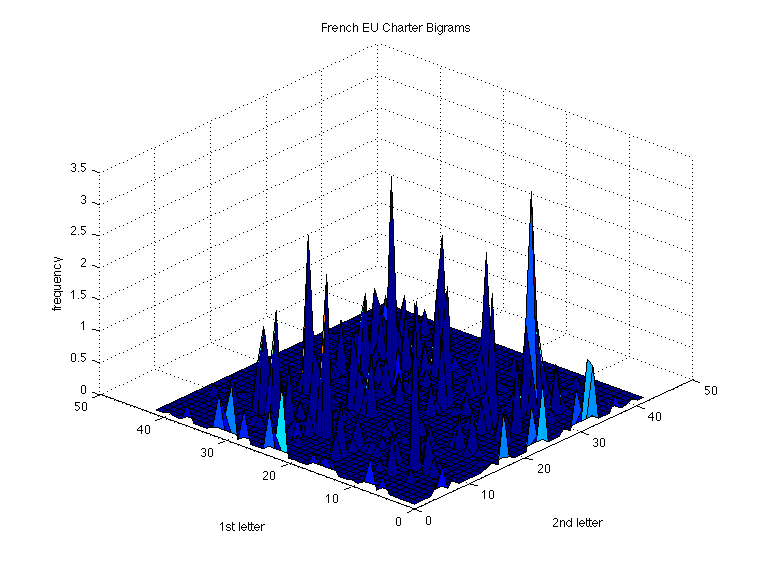}
	\end{center}
	\caption{Distribution of French bigrams} \label{distr_fr}
\end{figure}

\begin{figure}[h]
	\begin{center}
		\vspace{3mm}
		\includegraphics[width=0.6\textwidth]{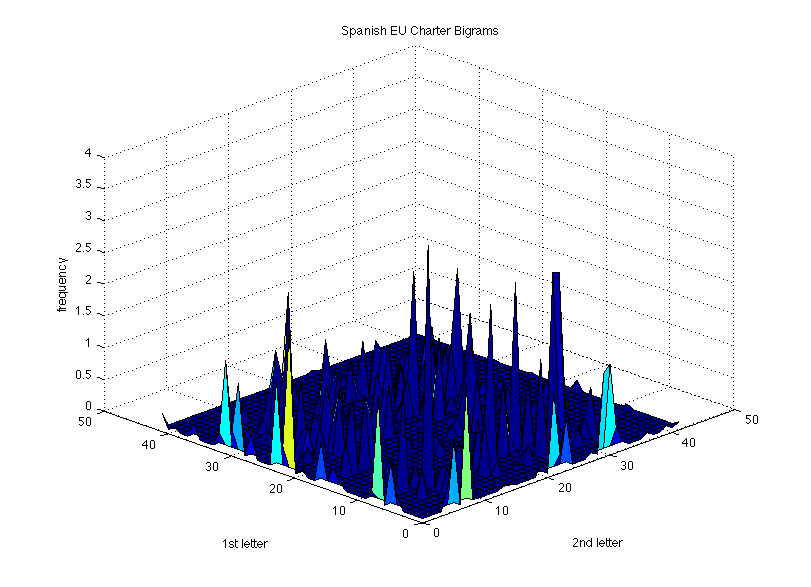}
	\end{center}
	\caption{Distribution of Spanish bigrams} \label{distr_es}
\end{figure}

The program was run using a series of files in English, French and Spanish. Initial frequency tables for analysis of individual languages were created using the EU Charter in each respective language \cite{eu}, producing the frequency distributions shown in Figures \ref{distr_en}, \ref{distr_fr}, and \ref{distr_es} for English, French, and Spanish, respectively.

These frequency graphs show that each language has a handful of extremely common bigrams (in addition to some which appear little or not at all). In English, this includes ``th'' and ``he'' with percentage frequencies of 2.9 and 2.8, respectively, along with ``on'' and ``ti'' also both above 2.5\%. This data is slightly skewed by the text used, though ``th'' and ``he'' are indeed the most common bigrams in English. A study conducted using a sample of 40,000 words \cite{freq} gave the two frequencies of 1.5 and 1.3, respectively, though the next most common bigrams in the sample text are not as common in the English language as a whole as their frequencies here would suggest. This is largely due to an inherent bias in the text, as words such as ``protec\underline{tion}'' or ``resp\underline{on}sibili\underline{ti}es'' appear frequently in the EU Charter.

French resulted in ``es'' and ``on'' as the most common bigrams, followed by ``de'' and ``le''. In Spanish, the most common bigram by far was ``de'', followed by ``cl'', ``en'', ``er'', and ``es''. Again, however, these likely suffer from slight biases due to the nature of the text.

The comparison threshold was initially determined using a series of randomized Wikipedia articles of considerable length in English and French. The same threshold was also used to compare English and Spanish texts and French and Spanish texts with continued high accuracy for texts of considerable length, showing that this method does not vary notably with different languages. The outputs of these tests are shown in Table \ref{samples}.

\begin{table}[h]
	\begin{center}
		\vspace{3mm}
		\begin{tabular}{|c|c|c|c|c|c|c|} \hline
			 & philosophy & encyclopedia & france (fr) & capitalism (fr) & jazz (es) & nyc (es) \\ \hline
			philosophy & \ldots & \ldots & \ldots & \ldots & \ldots & \ldots \\ \hline
			encyclopedia & 33.13 & \ldots & \ldots & \ldots & \ldots & \ldots \\ \hline
			france (fr) & 73.83 & 75.31 & \ldots & \ldots & \ldots & \ldots \\ \hline
			capitalism (fr) & 73.94 & 79.62 & 30.15 & \ldots & \ldots & \ldots \\ \hline
			jazz (es) & 67.68 & 69.56 & 64.76 & 67.76 & \ldots & \ldots \\ \hline
			nyc (es) & 71.76 & 73.42 & 66.41 & 70.46 & 28.95 & \ldots \\ \hline
		\end{tabular}
		\caption{Sample of sum differences} \label{samples}
	\end{center}
\end{table}

Finally, this method was tested with files of varying lengths. For the set of two English texts which were tested with decreasing word counts, the point at which this method was no longer accurate was between 400 and 200 words. For other files this is likely to vary, and could be lessened with further fine-tuning of the threshold. At a certain point, however, the difference values begin overlapping due to variation and bias from the words used in the text, making accuracy impossible.

\section{Syntax}

To analyze the syntax of a language, a ``recursive learning'' method is implemented using C++. Since the program would ideally require an absolute minimum of initial information, this method takes a small initial set of words and builds on this by alternately using known words to learn new sentence structures and using known sentence structures to learn new words as seen below.

\begin{center}
	\{cat, man, has\} $\rightarrow$ ``The man has a cat'' $\rightarrow$ ``The $x$ $y$ a $z$'' $\rightarrow$ ``The man wore a hat'' $\rightarrow$ \{cat, man, has, wore, hat\}
\end{center}

\subsection{Algorithm}

There are two main elements to understanding the recursive learning system, namely understanding both how the information is represented and the methods used to analyze new information.

\begin{figure}[htb]
	\begin{center}
		\vspace{3mm}
		\includegraphics[width=0.3\textwidth]{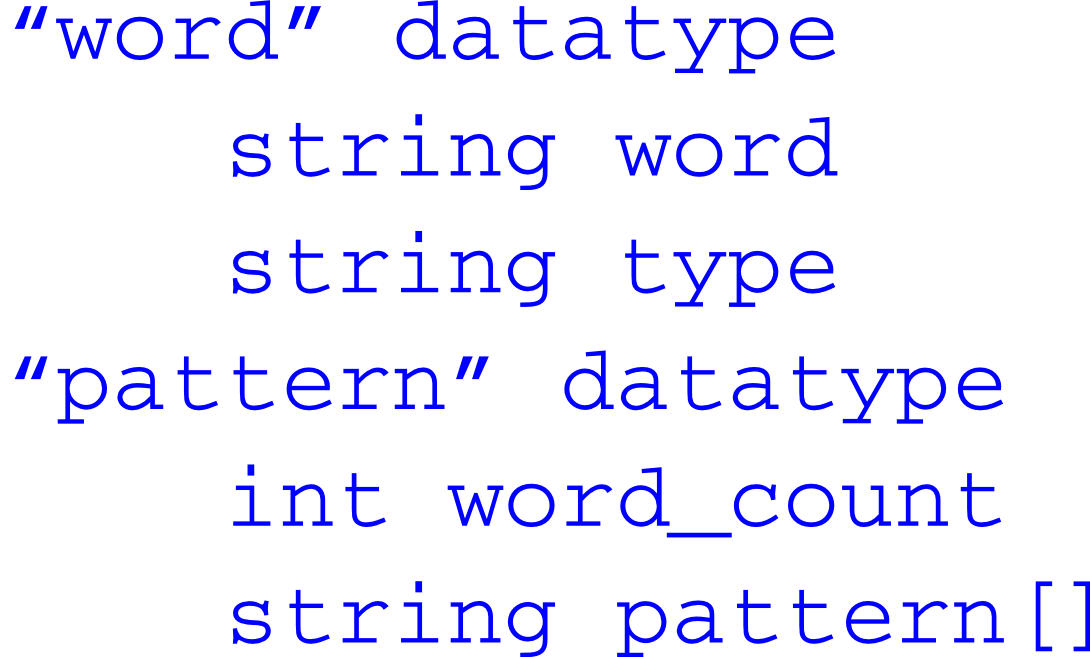}
	\end{center}
	\caption{Word and sentence pattern datatypes} \label{struct}
\end{figure}

The information this program gathers can be split into two pieces, information on the words themselves and information on valid sentence structures. For words, the program keeps track of the word itself and the word's type. Only two specific types are defined in the program, ``noun'' and ``verb'', and the actual meaning of each is defined by context. Any word which does not fit these definitions is defined relative to these definitions. For sentences, the program keeps track of the number of words in a given sentence pattern and the type of the word in each position (in an array). These structures are shown in Figure \ref{struct}.

\begin{figure}[htb]
	\begin{center}
		\vspace{3mm}
		\includegraphics[width=0.6\textwidth]{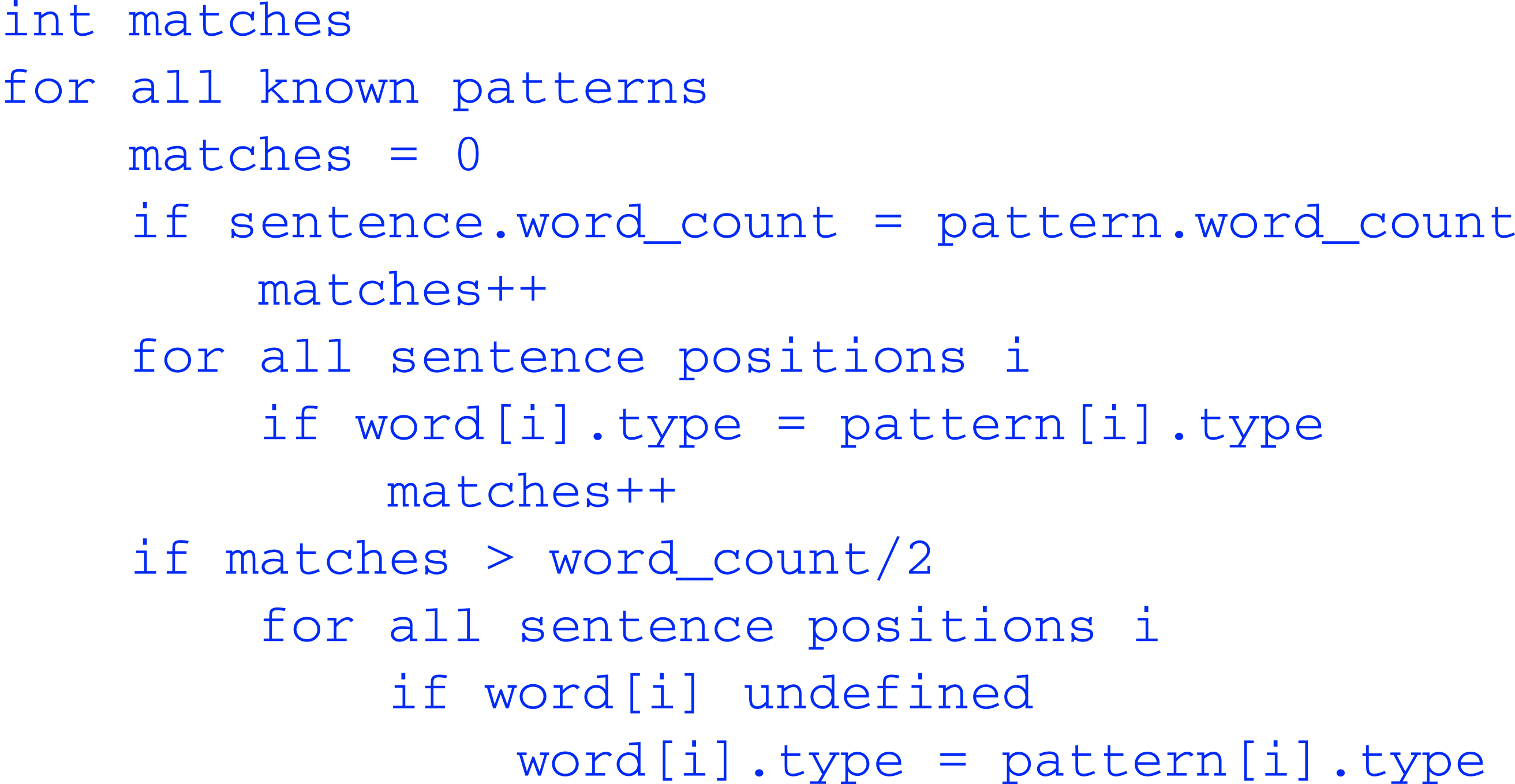}
	\end{center}
	\caption{Sentence structure method of learning} \label{patt}
\end{figure}

The methods used by the program to analyze new information can also be split into two pieces. Both methods require that some information is already known about the sentence in question, but are used in slightly different ways. The first method analyzes new words based on the structure of the sentence, by selecting the most applicable of existing patterns based on correspondence with known information. The program keeps track of how many ``matches'' there are between the sentence and the known pattern, taking the one with the most matches (if greater than half the word count) and using it to set unknown word types. This method is seen in Figure \ref{patt}

\begin{figure}[htb]
	\begin{center}
		\vspace{3mm}
		\includegraphics[width=0.6\textwidth]{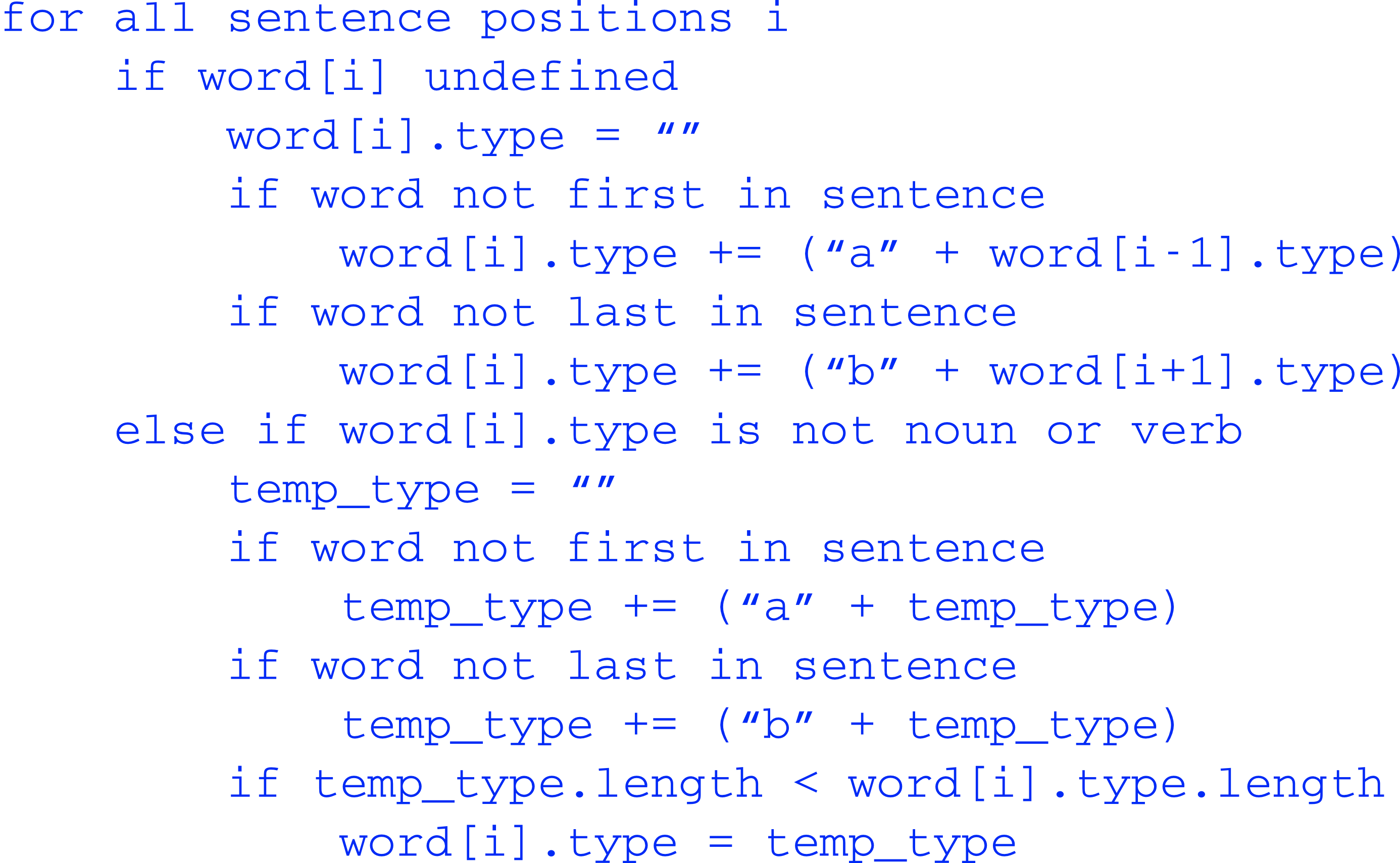}
	\end{center}
	\caption{Word-based method of learning} \label{cont}
\end{figure}

This method is particularly useful for learning new nouns and verbs, in situations where other words in the sentence are primarily known grammatical particles. The other method uses the surrounding words to define the type of any unknown words. The program notes the types of the words before and after the unknown word, and the type of the unknown word is then stated as ``a$<$type of previous word$>$ b$<$type of next word$>$''. For example, a word with type ``anoun bverb'' would be one which tends to come after nouns and before verbs. This type could also be analyzed and modified to give further insight into other types of words such as adjectives, as detailed in Section \ref{other}. The program can redefine known words if new information is found on their positioning, using the new definition if it is shorter, and thus more general, than the previous definition. This is seen in Figure \ref{cont}.

This method is best used when most of the nouns and verbs in a sentence are known, but other words exist which are not known. The program would be able to dynamically select between the two methods based on what information is already known. As a rule, the program would default to the first method initially as this is more likely to provide accurate results.

\subsection{Limitations}

With proper development this method could be used to learn many types of sentences. However, it still has several limitations. Although it is sufficient for simpler grammatical constructs, complicated words and patterns could cause confusion. In particular, words which have multiple meanings with different parts of speech would confuse the program tremendously, and at present it also has no way of connecting related words (such as plural or possessive forms of nouns).

In addition, this method requires carefully crafted training sentences. Although it can work with a minimum of initial information unlike most existing systems, it still has to learn the constructs somewhat sequentially and avoid having too many new concepts introduced at once. This could be partially remedied by implementing a method by which the program stores any sentences it does not yet have the tools to analyze to be recalled later, so that a more general text could be used as training material.

\subsection{Results}

As a proof of concept for these methods, a series of three training sentences were input and analyzed by the program to learn a handful of new words and concepts. Although the program begins with far less information than many existing programs, it nonetheless needs some initial input - in this case three words which will, together with ``a'' and ``the'', make up the initial sentence, as seen in Table \ref{init}.

\begin{table}[h]
	\begin{center}
		\vspace{3mm}
		\begin{tabular}{|c|c|} \hline
			\textbf{Word} & \textbf{Type} \\ \hline
			has & verb \\ \hline
			hat & noun \\ \hline
			man & noun \\ \hline
		\end{tabular}
		\caption{Initial program input} \label{init}
	\end{center}
\end{table}

The first sentence input into the program is ``the man has a hat,'' which is analyzed using the word-based second method. From here, two new words are learned - namely ``a'' and ``the'' which are defined based on the surrounding words. The sentence pattern is also catalogued, and the known information reads as in Table \ref{one}.

\begin{table}[h]
	\begin{center}
		\vspace{3mm}
		\begin{tabular}{|c|c|} \hline
			\textbf{Word} & \textbf{Type} \\ \hline
			has & verb \\ \hline
			hat & noun \\ \hline
			man & noun \\ \hline
			the & bnoun \\ \hline
			a & averb bnoun \\ \hline
		\end{tabular}
		\begin{tabular}{|c|} \hline
			5 bnoun|noun|verb|averb bnoun|noun \\ \hline
		\end{tabular}
		\caption{Program knowledge after run one} \label{one}
	\end{center}
\end{table}

At present, although ``the'' must only appear before a noun, ``a'' is assumed to require a preceding verb. To correctly define ``a'', the next sentence reads ``a man has the hat.'' While using the same words, the two known grammatical particles are reversed. The program redefines the word ``a'' with a more general definition (i.e. simply requires a succeeding noun), however the definition of ``the'' remains the same as the program determines that replacing the definition would add more constraints, which is counterproductive. The sentence pattern is again catalogued, and the known information reads as in Table \ref{two}.

\begin{table}[h]
	\begin{center}
		\vspace{3mm}
		\begin{tabular}{|c|c|} \hline
			\textbf{Word} & \textbf{Type} \\ \hline
			has & verb \\ \hline
			hat & noun \\ \hline
			man & noun \\ \hline
			the & bnoun \\ \hline
			a & bnoun \\ \hline
		\end{tabular}
		\begin{tabular}{|c|} \hline
			5 bnoun|noun|verb|averb bnoun|noun \\
			5 bnoun|noun|verb|bnoun|noun \\ \hline
		\end{tabular}
		\caption{Program knowledge after run two} \label{two}
	\end{center}
\end{table}

Although the two existing sentence patterns are functionally identical, and the first should actually be redefined as the second, both are kept to demonstrate the first method based on sentence patterns. For this, the sentence ``the dog ate a biscuit'' is used, having the same structure as existing sentences but a different set of nouns and verbs. Although some matches exist with the first pattern, the redefinition of ``a'' results in only two matches. Instead, the program finds this to be the same as the second sentence pattern, as the number of words matches as do the types of the words ``the'' and ``a''. Hence, the program defines the unknown words based on this pattern, resulting in the set of information shown in Table \ref{three}.

\begin{table}[h]
	\begin{center}
		\vspace{3mm}
		\begin{tabular}{|c|c|} \hline
			\textbf{Word} & \textbf{Type} \\ \hline
			has & verb \\ \hline
			hat & noun \\ \hline
			man & noun \\ \hline
			the & bnoun \\ \hline
			a & bnoun \\ \hline
			dog & noun \\ \hline
			ate & verb \\ \hline
			biscuit & noun \\ \hline
		\end{tabular}
		\begin{tabular}{|c|} \hline
			5 bnoun|noun|verb|averb bnoun|noun \\
			5 bnoun|noun|verb|bnoun|noun \\ \hline
		\end{tabular}
		\caption{Program knowledge after run three} \label{three}
	\end{center}
\end{table}

\section{Analysis}

In this project we have developed methods for allowing a computer to understand and learn both the morphology and the syntax of a language. Using novel techniques or applications, we have designed and implemented these methods and tested their capabilities for learning language.

Although the use of bigrams in language analysis is not a new idea, we have implemented it in a novel way by working to develop it as a defining quality and learning mechanism for natural language processing. The method proves very useful for understanding the morphology of a language, though only to a point. It is extremely effective when using a large enough sample text, but with smaller sample texts it is no longer able to accurately compare languages. Hence, although it creates a computationally effective ``definition'' of a language, its actual ability as a learning mechanism is limited. Used in tandem with other methods, the bigram method could prove extremely powerful.

The recursive learning method implemented for gaining an understanding of syntax proves very useful and has great potential to be developed further. Both of its submethods work with high accuracy for simple sentences, and hence it is able to develop a growing model of sentence construction. Even more particularly, it is able to do this with a very small amount of initial input and with methods which could be applied to many types of languages.

\subsection{Comparison to Existing Programs}

The use of bigrams to understand and analyze different parts of a given language has been studied and implemented substantially.  For example, there are programs that calculate bigram frequencies to evaluate a language's morphology.  However, unlike our program, none to date have utilized the differences in bigram frequencies between two languages to distinguish one language from the next.

The system in the program that was used for determining the parts of speech in a sentence has rarely been attempted, and when it has been used, only partially and in conjunction with other methods.  Most natural language processing programs have been designed to be as the efficient and effective as possible.  As a result, many use large banks of initial data, which the program then analyzes and uses for subsequent input.  As discussed previously, the most common and successful programs of this sort are statistical parsers. On the contrary, our program uses the recursive principle to acquire new vocabulary and forms of syntax for a given language, provided only a very small initial set of data.  In practice, our model only required one sentence with verb and noun indicated to determine the parts of speech of all other words, although not denoting them in linguistic terms (article, preposition, etc.), as well as learn new words and, in theory, to learn new sentence patterns.  Despite the extensive power of the recursive method, it has rarely been used in the history of natural language processing.  The results of our program illustrate the potential abilities of the recursive method that have not been seen previously.

\section{Extension}

The program as it stands now can learn and understand a range of language elements, including morphology and simple sentence patterns. However, there is still significant room for further exploration both by developing the techniques for learning syntax to allow for a more complete range of sentence patterns, and by developing methods for a computational understanding of semantics.

\subsection{Other Sentence Structures} \label{other}

In addition to existing sentence structures and constructs, the syntax program can learn other common constructs. Although some may be understandable at present, others may require some additions to the program to fully understand. Below are some examples of other simple sentence constructs and how the program would interpret them.

\emph{``The man has a blue hat.''}

Here, the only unknown word with present knowledge is ``blue''. The program at present would interpret it using the word-context method, resulting in a type of ``abnoun bnoun''. Continuing in this manner would quickly lead to complications, however, so the program could be extended to understand words based on the form of types which are not nouns or verbs. The word directly before blue has the type ``bnoun'', so ``blue'' would be interpreted in this manner as a noun, resulting in two nouns in a row. The program could additionally be edited to interpret two like words in a row as a ``noun phrase'' or ``verb phrase'', which would differentiate adjectives from nouns and adverbs from verbs in a way more intuitive to the computer.

\emph{``The hat is blue.''}

This sentence presents a different use of an adjective effectively in the place of a noun. Assuming ``is'' had been previously learned as a verb, this sentence would be more-or-less readily understandable as ``blue'' is still interpreted as related to nouns. This introduces another common sentence construct as well, namely that some nouns can appear directly after verbs without a ``bnoun'' word in between. Here, it may become worthwhile to add some indicator to new nouns about whether they can appear directly after a verb or not.

\emph{``The man has one hat.''}

This sentence would introduce numerical words, which would be readily understandable as ``bnoun'' words similar to ``a'' or ``the''. This is sufficient and accurate for most cases, though as the program expands into semantics this construct may require more specific definition.

\emph{``The man has four hats.''}

Here, the concept of plurality is introduced. This sentence would simply be another example of the above sentence with regards to syntax alone---the word ``hats'' would just be considered a new noun. However, this would likely be the most problematic concept regarding semantics. Without an external concept of meaning or some other indication, plurality would have to be learned simply by similarity at the word level. In many cases this would be sufficient, such as ``hats'', but some words do not follow standard plurality rules such as ``mice'' versus ``mouse''. Similar issues apply to verb tenses, though here the rules are even less standardized. This issue could be at least partially remedied by creating an artificial semblance of external understanding, though this would likely prove difficult as well.

Many other sentence constructs are built from these sorts of basic patterns. For example, another common sentence construct involves prepositions such as the sentence ``the man threw the hat in the trash''. Assuming prior knowledge of the nouns present, the program could interpret ``in'' as verb-like, appearing between two noun phrases. This is, again, a sufficient interpretation in most cases. A marker indicating what would effectively be two full constructs could be implemented as well. Embedded clauses would be interpreted similarly, where the program would find the pattern of the outside clause and then the pattern of the inside clause. An understanding of punctuation would make such sentences more easily interpretable.

Despite some issues, the program can, currently or with only minor modifications, understand many common sentence constructs using the two main forms of learning and still a fairly small amount of initial information. The program proves fairly versatile, though of course not at the full level of human understanding.

\subsection{Semantics}

Semantics is the meanings of words or sentences, not only determined by direct definition, but also by connotation and context.  To gain a computational understanding of semantics in a given language, without external representation for words, we have devised an associative approach.  For a noun, different adjectives and verbs can be used when employing the noun in a sentence.  The particular set of adjectives and verbs for that noun are representative of the nature of the object.  A second noun will have some shared verbs and adjectives, and the degree to which this is true will indicate the similarities of the two objects.  Conversely, a verb would be understood based on the different objects associated with it, especially the order of the nouns, i.e the subjects and objects used with that noun.  From a set of sentences, a threshold of similarity could be experimentally determined, which would result in the categorization of the different nouns and verbs.  The sequence of sentences could be as follows:

\begin{enumerate}
\item The man wears the big hat.
\item The man throws the small hat.\\
(``hat'' is known to be something that can be ``big'', ``small'', ``worn'', and ``thrown'')
\item The man throws the big ball.
\item The man bounces the small ball.\\
(``ball'' is like ``hat'' in having the ability to be ``big'' or ``small'' and can be ``thrown'', however it has not yet be found that it can be ``worn'', and ``hat'' has not be shown to be able to be ``bounced'')
\item The man wears the big shoes.
\item The man throws the small shoes.\\
(``shoes'' is then understood as the same type of object as ``hat'')
\item The man uses the telephone.
\item The man answers the telephone.
\end{enumerate}

\begin{figure}[htb]
	\begin{center}
		\vspace{3mm}
		\includegraphics[width=0.7\textwidth]{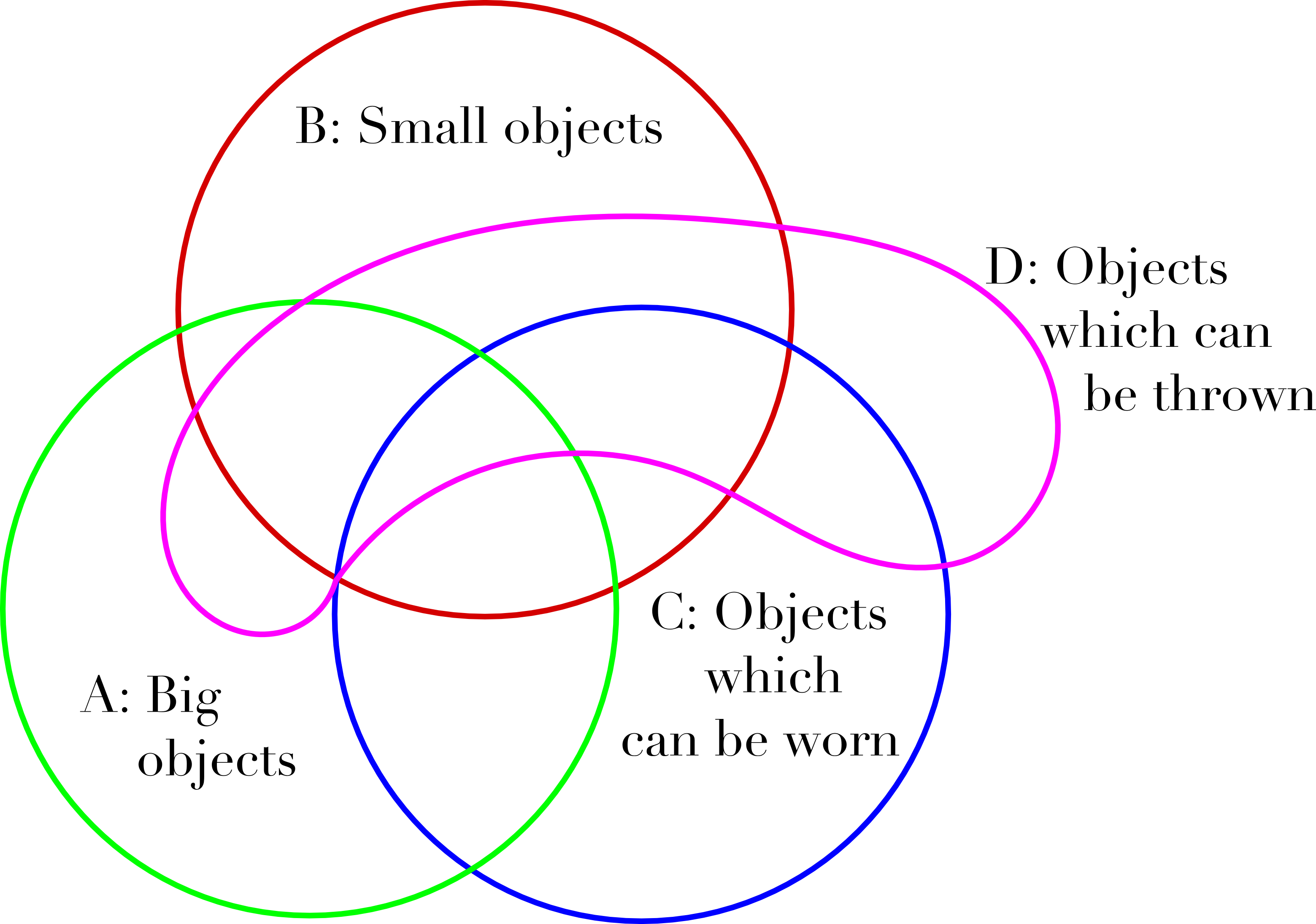}
	\end{center}
	\caption{Sample semantic web} \label{venn}
\end{figure}

Let us stop here because the fundamental logic of the sequence of sentences can now be seen.  Using these characteristics of nouns, a complex associative web would be formed, where objects have meaning based on their relation to a set of other objects, which also have meaning in relations to another set objects that the first may not. This web may be represented visually by a venn diagram similar to the one in Figure \ref{venn}.

For the verbs ``wear'', ``throw'', ``bounce'' etc., the program would interpret the above sentences as suggesting that only ``man'' can conduct these actions, but cannot be the recipient of these actions, while certain nouns can be the recipient of these actions.  Further, it should be noted that the training sentences do not require a strict order because a characteristic of a particular noun would be stored until another noun was found to have the same characteristic, and the two nouns would then share a link of the web.  Similarly, a verb would have tentative subjects and objects associated with it until more were found.  Nevertheless, this method is limited by the requirement of large amounts of sentences in order to gain any significant understanding of a given word.

To understand a concept such as time, which is essential for semantics and acknowledging tenses, the program would require some initial specification, as well as using a time indicator in all sentences referring to the past or present.  It would start with the conditional parameter that if a sentence has the word ``yesterday'', ``ago'', ``tomorrow'', ``later'', ``past'', or ``future'', then a time different from the current moment is being referenced, and the verb used is similar to another verb (the present form, which is the first form the program learns), but with a slightly different morphology.  Yet, a problem arises because ``last'' and ``next'' applied in ``last/next week'' or ``last/next month'' cannot be used strictly as indicators of time, as they can be used in other contexts, such as ``He ate the last cookie in the jar'' or ``He is next in line at the supermarket''.  Thus, in certain cases, time would present a difficulty for the semantic understanding of the program.  The same applies for the use of negation, as in the sentence ``The man is not big'', whereby ``not'', ``never'', ``none'', ``no longer'', and ``no more'' would have to be explained prior and would result in the noun or verb being categorized as different from other nouns and verbs with shared associative terms.

In addition, the associative method only allows for a definitional understanding of a given word, which can be substantially limited due to the possible changes in the meaning of a word as a result of its context. To allow a computer to have an understanding of context, a system could be implemented to keep track of previously learned information. Cognitive scientist Marvin Minsky suggests that a series of ``frames'' which represent generalized situations can be used to represent cognition computationally. These general frames could then be modified with situation-specific variables \cite{minsky}. This idea could prove useful in natural language processing to give a computer an understanding of context. For example, if a series of sentences read ``John is playing soccer. He kicked the ball,'' the program would be able to select a general frame which it could use to keep track of relevant variables, such as ``John'' being the subject of this action---hence linking this to the ``he'' in the next sentence.

Another issue that might arise is the issue of connotative meaning of words rather than merely denotative meaning. This is also related to the idea of symbolism, another element of language which can prove difficult for a computer to understand. Here, a method similar to the associative approach above could be implemented after the initial denotative associations were formed. Here, the training sentences would be ones using symbolism rather than literal ones as above. If a word is suddenly associated with a word which is not within the proper categorization, such as ``a heart of stone,'' it could be interpreted by the computer as a connotative association. This would allow the computer to examine characteristics related only to one or the other, hence gaining an understanding of the symbolic associations of words.

\section{Conclusion}

Using certain principles of language, we have designed a novel method by which a computer can gain an intuitive understanding of language rather than simply an artificial understanding. We have developed techniques by which a computer can learn and analyze the morphology of any given language, and hence understand differences between two languages. We have also developed a recursive learning system for understanding sentence patterns and constructs, which uses a minimum of initial information. At present, the program can interpret many basic sentences, and we have also provided possibilities and suggestions for extending the capabilities of the program. This approach is unique compared to common natural language processing systems because of this lack of need for significant initial input and its recursive design, and could have great potential in the field of natural language processing.

\begin{onehalfspacing}

\end{onehalfspacing}

\appendix

\section{Sample Texts}

Sample texts for the morphology-analysis system were primarily selected from randomized Wikipedia articles in the respective languages, and were selected primarily for length. The following sample texts were used for testing:

\vspace{5 mm}

{\footnotesize English EU Charter (available at \url{http://www.europarl.europa.eu/charter/pdf/text_en.pdf})

French EU Charter (available at \url{http://www.europarl.europa.eu/charter/pdf/text_fr.pdf})

Spanish EU Charter (available at \url{http://www.europarl.europa.eu/charter/pdf/text_es.pdf})

English Wikipedia article on ``Philosophy'' (available at \url{http://en.wikipedia.org/wiki/Philosophy})

English Wikipedia article on ``Encyclopedias'' (available at \url{http://en.wikipedia.org/wiki/Encyclopedia})

English Wikipedia article on ``Peace Dollars'' (available at \url{http://en.wikipedia.org/wiki/Peace_dollar})

English Wikipedia article on ``Buffalo Nickels'' (available at \url{http://en.wikipedia.org/wiki/Buffalo_nickel})

French Wikipedia article on ``France'' (available at \url{http://fr.wikipedia.org/wiki/France})

French Wikipedia article on ``Capitalism'' (available at \url{http://fr.wikipedia.org/wiki/Capitalisme})

French Wikipedia article on ``Karl Marx'' (available at \url{http://fr.wikipedia.org/wiki/Karl_Marx})

French Wikipedia article on ``Democracy'' (available at \url{http://fr.wikipedia.org/wiki/Démocratie})

Spanish Wikipedia article on ``Jazz'' (available at \url{http://es.wikipedia.org/wiki/Jazz})

Spanish Wikipedia article on ``New York City'' (available at \url{http://es.wikipedia.org/wiki/New_York_City})

Spanish Wikipedia article on ``the Sassanid Empire'' (available at \url{http://es.wikipedia.org/wiki/Imperio_sasánida})

Spanish Wikipedia article on ``the Crown of Castile'' (available at \url{http://es.wikipedia.org/wiki/Corona_de_Castilla})}

\section{Morphology Code}

\lstset{language=Java, tabsize=2, basicstyle=\footnotesize\ttfamily, showspaces=false, showstringspaces=false, extendedchars=false, escapeinside=''}

\begin{singlespacing}
\begin{lstlisting}
package ngrams;

import java.io.BufferedReader;
import java.io.File;
import java.io.FileWriter;
import java.io.IOException;
import java.io.InputStreamReader;
import java.io.PrintWriter;
import java.util.Scanner;

public class Ngrams {

	// Main program
	public static void main(String[] args) throws IOException {
		String file1 = "", file2 = "";
		float diff = 0; // Total frequency difference between two files
		char[] alpha = {'a','\`{a}','\^{a}','\"{a}','b','c','\c{c}','d','e','\`{e}','\'{e}','\^{e}','\"{e}','f','g','h','i','\^{i}','\"{i}','j','k','l','m','n','o','\^{o}',
			'p','q','r','s','t','u','\`{u}','\^{u}','\"{u}','v','w','x','y','\"{y}','z'}; // Valid characters
		System.out.print("Input First File Name: ");
		file1 = read(file1); // Reads text from file
		System.out.print("Input Second File Name: ");
		file2 = read(file2);
		float bi1[][] = count(file1, alpha); // Creates frequency table for file
		float bi2[][] = count(file2, alpha);
		// Calculates total frequency difference
		for (int i = 0; i < alpha.length; i++)
			for (int j = 0; j < alpha.length; j++)
				diff += Math.abs(bi1[i][j] - bi2[i][j]);
		System.out.println(diff);
		if (diff < 55) // Experimentally determined threshold
			System.out.println("Same Language");
		else
			System.out.println("Different Language");
	}

	// Create frequency tables
	public static float[][] count(String file,char[] alpha)
	throws IOException { // 2-d frequency table of all possible bigrams
		float bigram[][] = new float[alpha.length][alpha.length];
		for (int i = 0; i < alpha.length; i++)
			for (int j = 0; j < alpha.length; j++)
				bigram[i][j] = 0; // Initialize with frequency=0
		int a = alpha.length+1, b = alpha.length+1;
		float total = 0; // Total number of bigrams
		Scanner in = new Scanner(new File(file+".txt"));
		while(in.hasNext("\\S+")) { // Read file word by word
			String word = in.next("\\S+");
			word.toLowerCase();
			for (int k = 0; k < word.length()-1; k++)
			{ // For each pair of letters
				a = alpha.length+1;
				b = alpha.length+1;
				for (int m = 0; m < alpha.length; m++)
				{ // Locates each letter within alpha list
					if (word.charAt(k) == alpha[m])
						a = m;
					if (word.charAt(k+1) == alpha[m])
						b = m;
				}
				if (a < alpha.length && b < alpha.length)
				{ // Adds valid bigrams to frequency list
					bigram[a][b]++;
					total++;
				}
			}
		}
		// Create new file with frequency table
		PrintWriter out = new PrintWriter(new FileWriter(file+"_tab.csv"));
		for (int p = 0; p < alpha.length; p++) {
			for (int q = 0; q < alpha.length; q++) {
				bigram[p][q] = (bigram[p][q] / total) * 100;
					// Convert to decimal frequency
				out.print(bigram[p][q]);
				out.print(",");
			}
			out.print("\n");
		}
		out.close();
		return bigram;
	}

	// Read text from file
	public static String read(String str) {
		String str2 = "";
		BufferedReader read = new BufferedReader(new InputStreamReader(System.in));
		try {
			str2 = read.readLine();
		} catch (IOException ioe) {
			System.out.println("Error: Cannot Read Input\n");
			read(str);
		}
		return str2;
	}

}
\end{lstlisting}
\end{singlespacing}

\section{Syntax Code}

\lstset{language=C++, tabsize=2, basicstyle=\footnotesize\ttfamily, showspaces=false, showstringspaces=false}


\begin{singlespacing}
To begin, types must be created and initial input defined, and the sentence must be divided into its component words which are checked against known information:

\begin{lstlisting}
	// Type definitions
	struct word {
		string word;
		string type;
	} list[50];
	struct pattern {
		int wcount;
		string patt[10];
	} list2[10];
	// Initial input - First word must be "NULL" to avoid errors
	list[0].word = "NULL";
	list[0].type = "NULL";
	list[1].word = "has";
	list[1].type = "verb";
	list[2].word = "hat";
	list[2].type = "noun";
	list[3].word = "man";
	list[3].type = "noun";
	

	// Split sentence into component words
	for (i = 0; i < sen.length(); i++) {
		if (sen.at(i) == ' ')
			sen_count++;
	}
	for (i = 0; i < sen.length(); i++) {
		if (sen.at(i) == ' ') {
			words[pos] = sen.substr(temp,i-temp);
			temp = i+1;
			pos++;
		}
	}

	// Find words in "list" array
	num2 = num;
	for (i = 0; i < sen_count; i++) { // For each word
		for (j = 0; j < num; j++) {
			if (words[i] == list[j].word) {
				inarray[i] = j; // For known words, track position
			}
		}
		if (inarray[i] == 0) {
			inarray[i] = num2; // For unknown words, set new position
			num2++;
		}
	}
\end{lstlisting}

\vspace{5 mm}

\noindent
Unknown words are then defined using either the word-context method:

\begin{lstlisting}
	// For words not in "list" array, define

	// Define unknown words by surrounding words
	for (i = 0; i < sen_count; i++) { // For each word
		if (inarray[i] >= num)
		{ // If unknown, position will be greater than num (number of words)
			list[inarray[i]].word = words[i];
			list[inarray[i]].type = "";
			if (i > 0) // If not the first word in the sentence
				list[inarray[i]].type = 
				list[inarray[i]].type + "a" + list[inarray[i-1]].type + " ";
			if (i < sen_count-1) // If not the last word in the sentence
				list[inarray[i]].type = 
				list[inarray[i]].type + "b" + list[inarray[i+1]].type + " ";
		}
		else if (list[inarray[i]].type != "noun" &&
		list[inarray[i]].type != "verb")
		{ // If known, but not a noun or verb, checks if should be redefined
			type_temp = "";
			if (i > 0)
				type_temp = type_temp + "a" + list[inarray[i-1]].type + " ";
			if (i < sen_count-1)
				type_temp = type_temp + "b" + list[inarray[i+1]].type + " ";
			if (type_temp.length() <= list[inarray[i]].type.length())
				// The shorter one is the type
				list[inarray[i]].type = type_temp;
		}
	}
	num = num2; // Reset known number of words

	// Create new sentence pattern
	list2[runs].wcount = sen_count;
	for (i = 0; i < sen_count; i++) { // For each word
		list2[runs].patt[i] = list[inarray[i]].type;
	}
\end{lstlisting}

\vspace{5 mm}

\noindent
Or by the sentence pattern method:

\begin{lstlisting}	
	// Define unknown words by sentence structure
	for (j = 0; j < 2; j++) { // For each existing pattern
		matches = 0;
		for (i = 0; i < sen_count; i++) { // For each word
			if (inarray[i] < num)
				if (list[inarray[i]].type == list2[j].patt[i])
					matches++;
		}
		if (sen_count == list2[j].wcount) // If word counts match
			matches++;
		if (matches > sen_count/2)
		{ // If enough matches exist, unknown words are defined
			for (i = 0; i < sen_count; i++) {
				if (inarray[i] >= num) {
					list[inarray[i]].word = words[i];
					list[inarray[i]].type = list2[j].patt[i];
				}
			}
		}
	}
	num = num2; // Reset known number of words
\end{lstlisting}
\end{singlespacing}


\begin{thebibliography}{99}

\bibitem{turing} E. Reingold. The Turing Test. University of Toronto Department of Psychology. Available at \url{http://www.psych.utoronto.ca/users/reingold/courses/ai/turing.html}.

\bibitem{siri} Learn More About Siri. Apple Inc. Available at \url{http://www.apple.com/iphone/features/siri-faq.html}.

\bibitem{lingpipe} LingPipe. Alias-I. Available at \url{http://alias-i.com/lingpipe/}.

\bibitem{stanford} D. Klein and C.D. Manning. Accurate Unlexicalized Parsing. In: \emph{Proceedings of the 41st Annual Meeting on Association for Computational Linguistics}, Vol. 1 (2003), 423--430.

\bibitem{berkeley} M. Bansal and D. Klein. Simple, Accurate Parsing with an All-Fragments Grammar. In: \emph{Proceedings of the 48th Annual Meeting on Association for Computational Linguistics} (2010).

\bibitem{stats} E. Charniak. Statistical Parsing with a Context-Free Grammar and Word Statistics. In: \emph{Proceedings of the Fourteenth National Conference on Artificial Intelligence}. AAAI Press/MIT Press, Menlo Park, CA (1997), 598--603.

\bibitem{elang} C.M. Powell. From E-Language to I-Language: Foundations of a Pre-Processor for the Construction Integration Model. Oxford Brookes University (2005).

\bibitem{cipher} R. Morelli. The Vigenere Cipher. Trinity College Department of Computer Science. Available at \url{http://www.cs.trincoll.edu/~crypto/historical/vigenere.html}.

\bibitem{markov} H.H. Chen and Y.S. Lee. Approximate N-Gram Markov Model Natural Language Generation. National Taiwan University Department of Computer Science and Information Engineering (1994).

\bibitem{poem} Dr. Seuss. \emph{Fox in Socks}. Available at \url{http://ai.eecs.umich.edu/people/dreeves/Fox-In-Socks.txt}.

\bibitem{recursion} M.H. Christiansen and N. Chater. Constituency and Recursion in Language. In: M.A. Arbib. \emph{The Handbook of Brain Theory and Neural Networks}. 2nd ed. MIT Press, Cambridge, MA (2003), 267--271.

\bibitem{child} B.C. Lust. \emph{Child Language: Acquisition and Growth}. Cambridge University Press, Cambridge, UK (2006).

\bibitem{eu} The Charter of Fundamental Rights of the European Union. European Parliament. Available at \url{http://www.europarl.europa.eu/charter/default_en.htm}.

\bibitem{freq} Digraph Frequency. Cornell University Department of Mathematics. Available at \url{http://www.math.cornell.edu/~mec/2003-2004/cryptography/subs/digraphs.html}.

\bibitem{minsky} M. Minsky. A Framework for Representing Knowledge. In: J. Haugeland, editor. \emph{Mind Design}. The MIT Press, Cambridge, MA (1981), 95--128.

\end{thebibliography}
\end{document}